# Automated co-design of high-performance thermodynamic cycles via graph-based hierarchical reinforcement learning

Wenqing Li[a], Xu Feng[a], Peixue Jiang[a,b], Yinhai Zhu[a,b,*]

a: Key Laboratory for Thermal Science and Power Engineering of Ministry of Education, Beijing Key Laboratory of $CO_2$ Utilization and Reduction Technology, Department of Energy and Power Engineering, Tsinghua University, Beijing, 100084, China

b: Shanxi Research Institute for Clear Energy Tsinghua University, Taiyuan, 030032, China

[*]**Corresponding Author**: Yinhai Zhu; Email: yinhai.zhu@tsinghua.edu.cn

**Abstract**

Thermodynamic cycles are pivotal in determining the efficacy of energy conversion systems. Traditional design methodologies, which rely on expert knowledge or exhaustive enumeration, are inefficient and lack scalability, thereby constraining the discovery of high-performance cycles. In this study, we introduce a graph-based hierarchical reinforcement learning approach for the co-design of structure parameters in thermodynamic cycles. These cycles are encoded as graphs, with components and connections depicted as nodes and edges, adhering to grammatical constraints. A deep learning-based thermophysical surrogate facilitates stable graph decoding and the simultaneous resolution of global parameters. Building on this foundation, we develop a hierarchical reinforcement learning framework wherein a high-level manager explores structural evolution and proposes candidate configurations, whereas a low-level worker optimizes parameters and provides performance rewards to steer the search towards high-performance regions. By integrating graph representation, thermophysical surrogate, and manager-worker learning, this method establishes a fully automated pipeline for encoding, decoding, and co-optimization. Using heat pump and heat engine cycles as case studies, the results demonstrate that the proposed method not only replicates classical cycle configurations but also identifies 18 and 21 novel heat pump and heat engine cycles, respectively. Relative to classical cycles, the novel configurations exhibit performance improvements of 4.6% and 133.3%, respectively, surpassing the traditional designs. This method effectively balances efficiency with broad applicability, providing a practical and scalable intelligent alternative to expert-driven thermodynamic cycle design.

# 1 Introduction

As the global energy landscape shifts toward deep decarbonization, the efficient conversion and utilization of energy within complex constraints has emerged as a central challenge in energy science and engineering[1, 2, 3]. Energy demand continues to rise across key sectors such as industry, buildings, and transportation[4]. Achieving carbon neutrality necessitates the development of thermal systems that are efficient, flexible, and scalable[5, 6]. Consequently, considerable research efforts have concentrated on optimizing performance and integrating representative thermodynamic cycles, including Brayton cycles[7, 8, 9], organic Rankine cycles[10, 11, 12], and the refrigeration heat pump cycles[13, 14]. Systematic investigations have addressed cycle configurations, working fluid selection, and operational strategies to enhance both efficiency and economic viability. According to the second law of thermodynamics, irreversible losses constrain the efficiency of real cycles, with the Carnot cycle efficiency serving as a fundamental theoretical limit for optimization[15, 16].

Nonetheless, prevailing studies predominantly adopt a "fixed structure plus parameter optimization" paradigm, wherein the cycle topology is predefined based on expert knowledge, followed by parameter tuning to balance thermodynamic and economic objectives. While this approach facilitates the identification of superior designs through comparative analysis, it remains heavily reliant on expert assumptions, thereby limiting the exploration of the design space and constraining the discovery of novel cycle configurations.

To address these limitations, recent research has explored alternative methodologies that optimize cycle structures from fundamental principles, including heat exchanger network (HEN) methods, superstructure approaches, graph theory techniques, and thermodynamic process synthesis strategies. The HEN method mathematically integrates all heat and cold streams within a system to minimize the number of heat exchangers and optimize network topology[17]. Superstructure methods represent component or technology selection as binary variables, transforming structural optimization into mixed-integer nonlinear programming (MINLP) problems[18]. Graph theory abstracts cycles as directed graphs, with components as nodes and pipelines as directed edges, enabling topological optimization via node and edge combinations[19, 20, 21]. Thermodynamic process synthesis constructs cycles by assembling sequences of fundamental processes, offering a more elemental depiction of energy conversion mechanisms[22].

These approaches transition thermodynamic cycle design from expert-driven specifications to algorithmic generation, opening new avenues for optimization. However, they remain constrained by computational inefficiencies and limited optimization performance, falling short of autonomous learning and cycle optimization. Overall, their algorithmic intelligence is insufficient.

Reinforcement learning (RL)[23], a subset of machine learning, enables agents to iteratively learn optimal policies through environmental interactions[24]. Recent advances in deep reinforcement learning (DRL)[25] have achieved notable success in continuous control, combinatorial optimization, and scientific discovery domains. Notable examples include AlphaGo and AlphaStar[26, 27], which demonstrate RL's capacity to meet or exceed expert-level performance in complex decision-making tasks. These findings indicate that RL functions not merely as an optimization tool but also as a means to uncover latent patterns within intricate decision processes.

Within energy systems, RL has exhibited substantial potential across diverse applications such as renewable energy integration, power grid regulation, and dynamic energy system management, thereby transforming conventional energy utilization paradigms toward greater intelligence[28, 29, 30, 31, 32].

Nevertheless, direct application of RL to thermodynamic cycle optimization presents significant challenges. Firstly, thermodynamic cycle design constitutes a combined optimization problem integrating discrete structural decisions with continuous parameter tuning. This necessitates simultaneous optimization of cycle topology and operating parameters, complicating the training of monolithic RL models. Secondly, the environmental reward signal is contingent on system performance, which can only be evaluated post continuous parameter optimization, resulting in delayed and computationally expensive feedback[33].

Hierarchical reinforcement learning (HRL)[34, 35, 36] addresses these challenges by decomposing complex, multi-objective tasks into simpler subtasks, thereby reducing complexity and enhancing policy learning efficiency. HRL thus offers a promising framework for thermodynamic cycle optimization.

Accordingly, this study proposes a graph-based hierarchical reinforcement learning approach for the co-design of thermodynamic cycle structure and parameters. Cycles are encoded as graphs, representing components and connections as nodes and edges subject to grammatical constraints. A deep learning thermophysical surrogate model facilitates stable graph decoding and simultaneous global parameter solution. Building upon this, a hierarchical reinforcement learning framework integrates graph representation, thermophysical surrogates, and Manager-Worker learning to establish a fully automated pipeline for encoding, decoding, and co-optimization. Using heat pump and heat engine cycles as case studies, the results demonstrate that the proposed method not only replicates classical cycle configurations but also identifies novel high-performance cycle architectures that surpass conventional designs, thereby balancing effectiveness and generality. Figure 1 presents the overall framework of the proposed methodology, and Figure 2 further illustrates its typical application scenarios in heat pump systems, heat engine systems, and integrated energy systems.

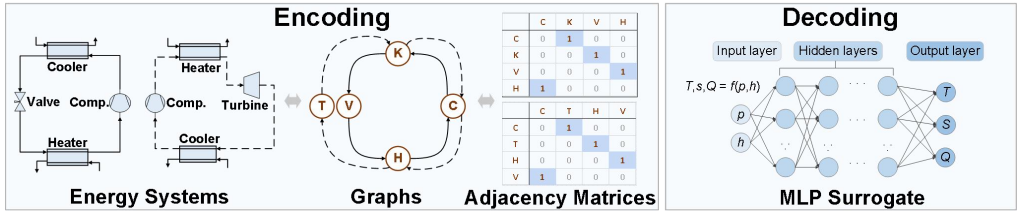

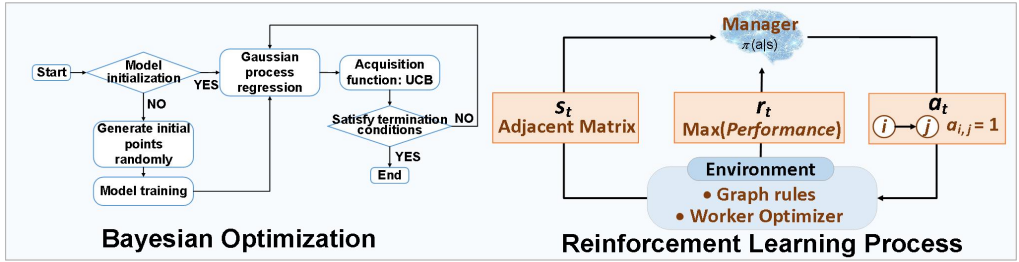

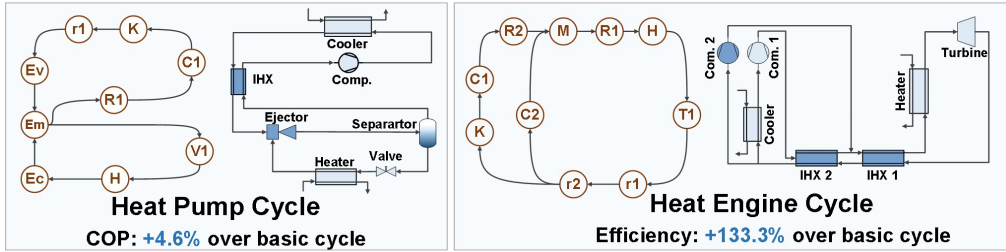

Figure 1 Overview of the graph-based, hierarchical reinforcement learning framework for automated structure–parameter co-design of thermodynamic cycles.

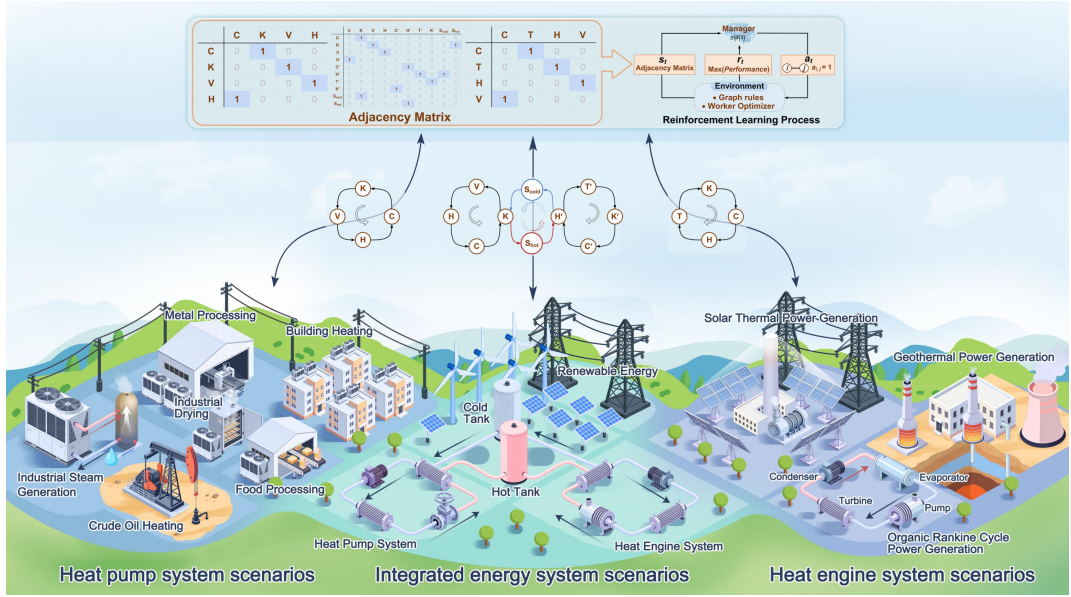

Figure 2 Typical application scenarios of the proposed method in heat pump, heat engine, and integrated energy systems. Heat pump system scenarios: industrial steam generation, metal processing, industrial drying, crude oil heating, food processing, building heating. Heat engine system scenarios: concentrated solar thermal power generation, Organic Rankine Cycle (ORC), geothermal power generation. Integrated energy system scenarios: renewable energy, power grid, energy storage.

## 2 Results

### 2.1 Graph-based thermodynamic cycle encoding

Graph-based encoding employs graph theory to represent thermodynamic cycles as mathematical objects interpretable by machine learning algorithms, constituting a prerequisite for automated cycle optimization. Figure 1 illustrates the proposed methodology, wherein the encoding process constructs the graph (G) and adjacency matrix (A) exemplified by heat pump and heat engine systems. The introduction of new components extends the adjacency matrix accordingly, enabling cycle optimization to be conceptualized as sequential moves on the adjacency matrix. Each move activates an edge ($a_{i,j}$ =1), signifying a new connection from component $i$ to $j$. Thus, optimizing the cycle reduces to identifying an optimal connectivity path within the graph space.

Analogous to board games, graph construction adheres to rules governing edge activation and final validity checks. Figure 3a-g detail these edge activation rules, encompassing general rules and special component rules. For instance, the internal heat exchanger (IHX) is modeled as two coupled nodes, R and r, representing heat transfer on low- and high-temperature sides, respectively. The ejector (E), a compound pressure component, decomposes into three coupled nodes (Ec, Ev, Em) corresponding to distinct thermodynamic effects. The gas-liquid separator

(S) is represented by three nodes (S, SV, SL), with SV and SL outputting gas and liquid phases, respectively. To resolve ambiguity in edges converging on a single node, a merging node (M) is introduced upstream of nodes with multiple inlets. Except for M, all nodes possess a single inlet.

Edge activation rules ensure local connection validity, while final validity checks assess global compliance with physical constraints, including connection, pressure, energy, parallelism, and heat transfer validity. Figure 3h-l exemplify violations of these constraints, underscoring their importance in guaranteeing physically feasible thermodynamic cycles.

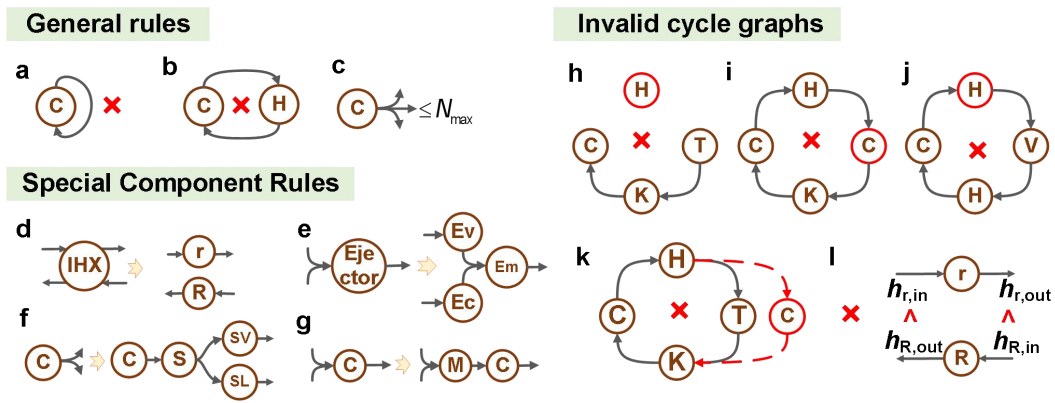

Figure 3 Graph construction rules.

General rules: a, $a_{i,i} \neq 1$; b, $a_{i,j} + a_{j,i} \leq 1$; c, $\sum_j a_{i,j} \leq N_{max}$.

Special component rules:

d, Internal Heat Exchanger (IHX), where R denotes the heat transfer process on the low-temperature side, and r denotes that on the high-temperature side;

e, Ejector (E), with Ev representing pressure reduction and enthalpy increase for the primary flow, Ec representing pressure increase and enthalpy reduction for the secondary flow, and Em representing the mixing process;

f, Gas-Liquid Separator (S), where the SV node outputs the gas phase and the SL node outputs the liquid phase.

G, M, merging node.

Invalid cycle graphs:

h, violation of connection validity, where all activated nodes must lie within a closed loop;

i, violation of pressure validity, requiring each loop to contain at least one pressure-increasing component and one pressure-decreasing component;

j, violation of energy validity, necessitating each loop to contain at least one enthalpy-increasing component and one enthalpy-decreasing component;

k, violation of parallelism validity, where the attributes of parallel branches must be consistent, and $\Delta P$ and $\Delta h$ directions must be the same;

l, violation of heat transfer validity, where heat exchange components must satisfy the pinch point temperature difference constraint to avoid temperature crossover.

**2.2 Physics-informed graph decoding**

Graph decoding integrates the graph structure with physical models to solve for inlet and outlet states of components and overall system performance. Traditional sequential point-by-point solution methods require manual determination of solution sequences, unsuitable for uncertain graph-derived systems. This study adopts a simultaneous solution approach for all component state points.

Accurate calculation of thermophysical properties for working fluids such as $CO_2$ typically involves solving complex state equations using databases like NIST REFPROP or CoolProp[37, 38]. However, simultaneous solution across all components often leads to numerical instability and convergence failures due to out-of-bound parameters. To mitigate this, a deep learning surrogate model[39] replaces conventional numerical solvers, providing stable, continuous, and differentiable property predictions with relative errors below 1% (Table S1), thus ensuring numerical robustness.

**2.3 Graph optimization based on hierarchical reinforcement learning**

The optimization space for system graphs is characterized by hybrid discrete-continuous variables with variable dimensionality: discrete adjacency matrix topology and continuous node operating parameters dependent on topology. This mixed discrete–continuous optimization with variable parameter spaces challenges conventional single-level optimization strategies.

To address this, a Manager-Worker hierarchical reinforcement learning framework is developed. The high-level Manager agent generates and optimizes graph structures via discrete actions, while the low-level Worker optimizes continuous parameters for a fixed structure and returns performance-based rewards.

Figure 1's reinforcement learning section depicts the Manager-environment interaction, where the environment comprises graph rules and the Worker parameter optimization module. At each step, the Manager activates an edge ($a_{i,j}=1$) based on the current graph state; the environment decodes the graph, optimizes parameters, evaluates performance, and feeds back rewards and updated adjacency matrices.

However, two principal challenges arise: (1) performance metrics (e.g., COP, thermal efficiency) are available only after cycle completion, causing delayed and sparse rewards; (2) the high-dimensional discrete space predisposes the Manager's policy to local optima, leading to instability and loss of high-performing structures.

To overcome these, a joint Manager-Worker training strategy incorporating performance feedback propagation, elite cycle memory, and staged training is proposed, enhancing exploration efficiency and convergence stability. Detailed methodology is presented in the Methods section.

**2.4 Automated co-design of high-performance heat pump cycles**

The air-source transcritical $CO_2$ heat pump cycle serves as a case study validating the proposed Manager-Worker HRL method. Figure 4a shows that the HRL agent's probability of generating valid cycles increases with training, stabilizing at 93.50% after 5,000 episodes, producing 4,675 valid and 325 invalid cycles. In contrast, random search yields a negligible valid cycle generation probability (~0.02%), indicating poor efficiency.

Figure 5 displays all 22 valid heat pump cycle structures discovered under graph rules and component constraints, classified by structural complexity and fluid split location into simple cycles (Nos. 1 - 3), complex cycles with high-/medium-pressure fluid splits (Nos. 4 - 11), and complex cycles with low-pressure fluid splits (Nos. 12 - 22). Notably, cycles 1, 2, 3, and 5 correspond to classical or expert-designed configurations, while the remainder are novel discoveries.

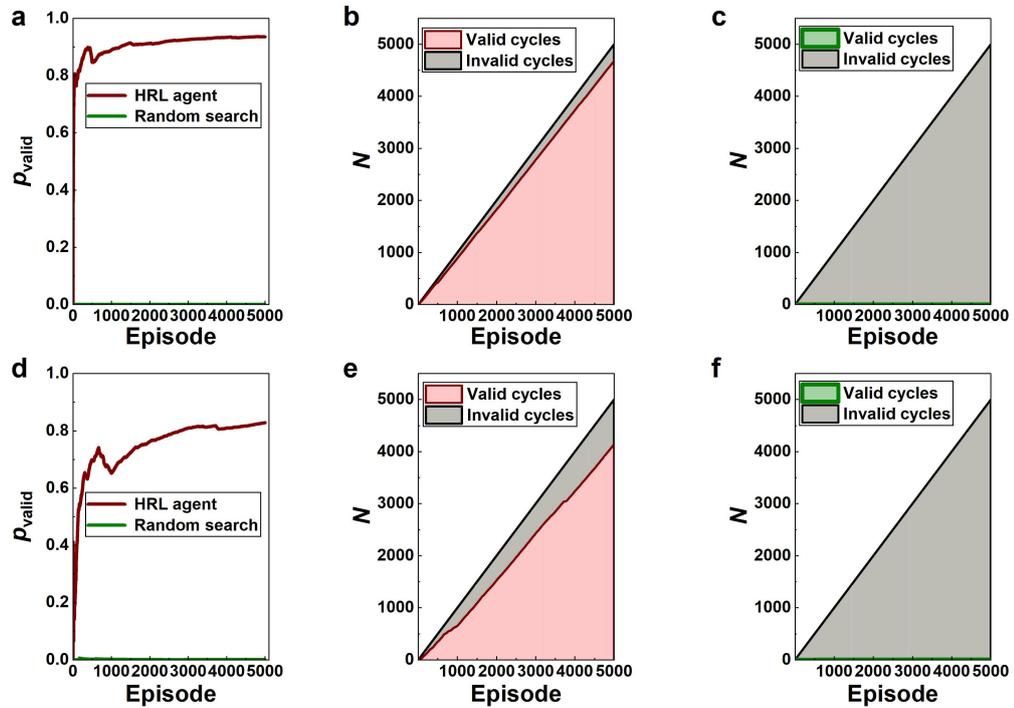

Figure 4 Comparison of valid cycle generation by the HRL agent and random search. a, Probability of generating valid heat pump cycles; b, Number of valid and invalid heat pump cycles generated by the HRL agent; c, Number of valid and invalid heat pump cycles generated by random search; d, Probability of generating valid heat engine cycles; e, Number of valid and invalid heat engine cycles generated by the HRL agent; f, Number of valid and invalid heat engine cycles generated by random search. Episode denotes the training iteration, $p_{valid}$ the probability of generating valid cycles, and $N$ the number of cycles.

Performance evaluation (Figure 6a-c) reveals significant variation in maximum COP

across structures under three operating conditions. Under case 1, cycles 3, 5, 6, and 7 achieve peak COP of 5.256, a 4.6% improvement over baseline cycle 1. Under higher heating temperatures (cases 2 and 3), cycle rankings shift, reflecting strong coupling between structure and operating conditions. Cycle 5 consistently attains the highest COP across all conditions, demonstrating robustness.

These findings indicate that the Manager-Worker HRL framework not only optimizes cycles under fixed conditions but also adaptively modifies structural preferences in response to operating environment changes, driven directly by COP optimization rather than memorization.

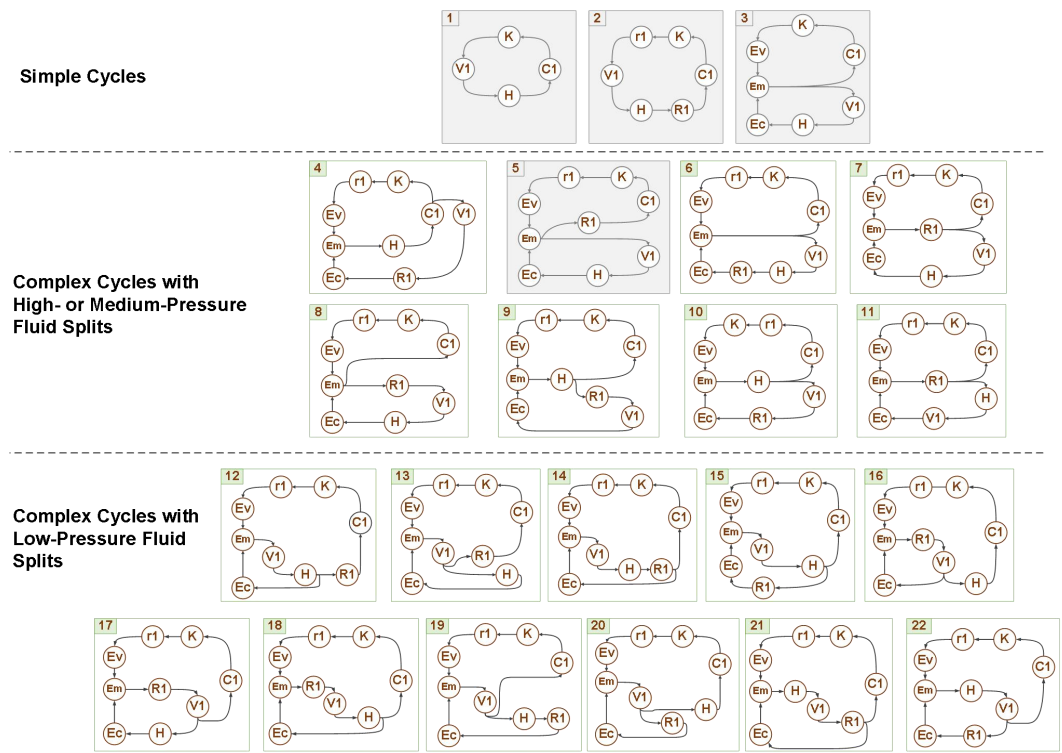

Figure 5 All valid heat pump cycles discovered by the HRL agent. The maximum number of key components-including compressor, gas cooler, evaporator, expansion valve, regenerator, and ejector-is limited to one, and COP is used as the performance metric. Detailed physical modeling of each component is provided in our previous study[14]. Cycles 1, 2, 3, and 5 correspond to expert-designed or classical configurations and are highlighted with a light grey background, whereas the remaining 18 cycles are newly discovered by the agent and are marked with green borders.

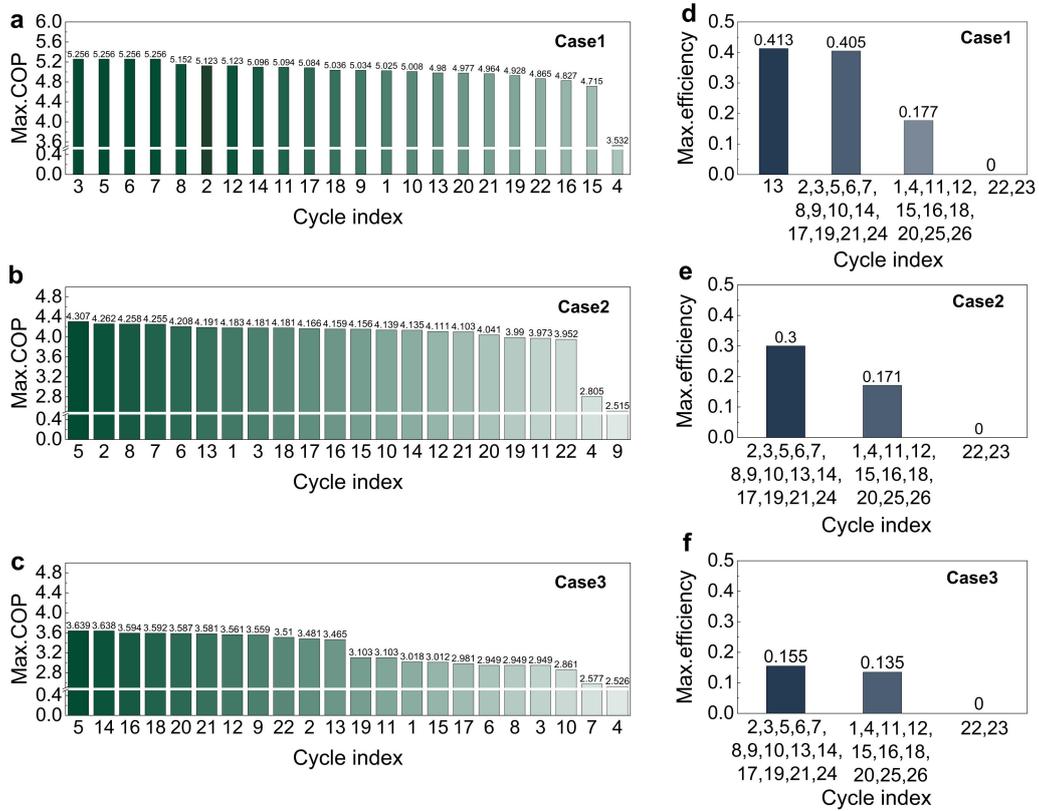

Figure 6 Cycle index versus optimal performance.

a, Heat pump cycles-maximum COP, case 1 (air source temperature: 20 °C; gas cooler outlet water temperature: 60 °C);

b, Heat pump cycles-maximum COP, case 2 (air source temperature: 20 °C; gas cooler outlet water temperature: 80 °C);

c, Heat pump cycles-maximum COP, case 3 (air source temperature: 20 °C; gas cooler outlet water temperature: 100 °C);

d, Heat engine cycles-maximum efficiency, case 1 (heat source temperature: 600 °C; heat sink temperature: 30 °C);

e, Heat engine cycles-maximum efficiency, case 2 (heat source temperature: 400 °C; heat sink temperature: 30 °C);

f, Heat engine cycles-maximum efficiency, case 3 (heat source temperature: 200 °C; heat sink temperature: 30 °C).

## 2.5 Automated co-design of high-performance heat engine cycles

The framework's generality is further assessed using the supercritical $CO_2$ Brayton cycle. Figure 4d shows the HRL agent's valid cycle generation probability rising to 82.82% over 5,000 episodes, yielding 4,141 valid cycles, whereas random search remains ineffective (<0.06%).

Figure 7 presents all 26 valid heat engine cycles discovered, categorized by compression configuration into single-stage (Nos. 1-4), two-stage (Nos. 5-12), and parallel compression (Nos. 13-26). Cycles 1, 2, 3, 8, and 13 correspond to known configurations; others are newly identified. Efficiency evaluation (Figure 6d-f) indicates cycle 13 achieves the highest thermal efficiency (0.413) under case 1, representing a 133.3% improvement over baseline.

Figure 6d~f illustrate the maximum efficiencies of these cycles after Worker optimization. Under case 1, cycle 13 achieves the best performance, reaching a maximum thermal efficiency of 0.413, representing a 133.3% improvement over the baseline (No. 1).

Notably, some cycles converge to identical efficiencies after optimization due to component redundancy under maximum-efficiency conditions, with components like IHX or compressors becoming effectively inactive. As a result, the corresponding cycle structures degenerate into simpler baseline or regenerative cycles.

Unlike heat pump cycles, heat engine cycle performance rankings remain largely stable across varying heat source temperatures, reflecting invariant underlying energy conversion mechanisms. This underscores the framework's reliability in identifying high-potential structures under diverse conditions.

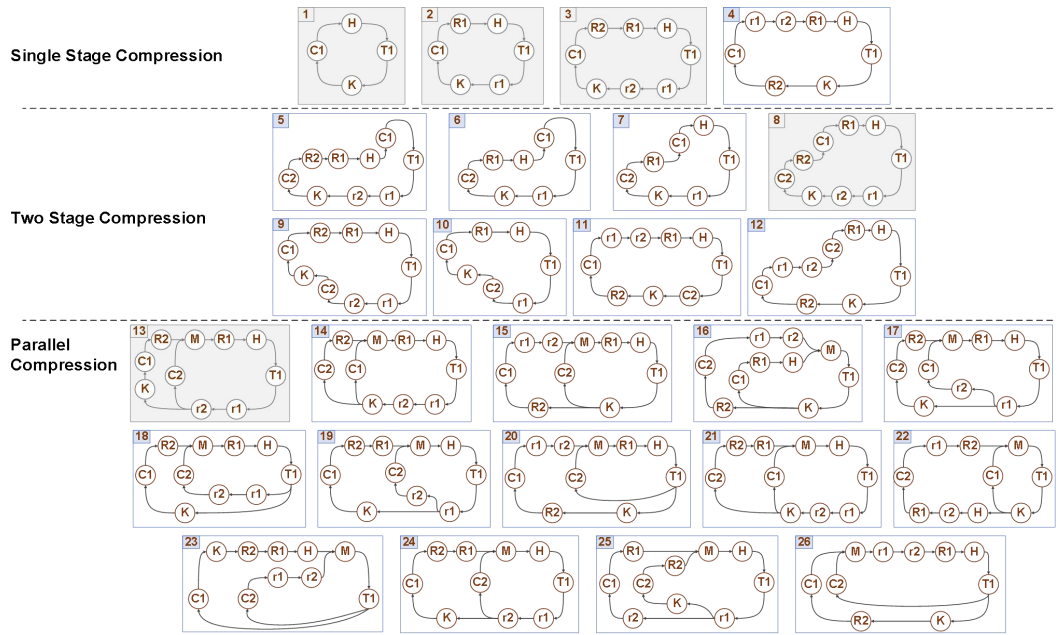

Figure 7 All valid heat engine cycles discovered by the HRL agent. The maximum number of turbines, heaters, and coolers is limited to one each, while up to two compressors and IHXs are permitted. Thermal efficiency $\eta_t$ serves as the performance metric. Cycles 1, 2, 3, 8, and 13 correspond to expert-designed or classical configurations and are highlighted with a light grey background, while the remaining 21 cycles are newly discovered by the agent and are marked with blue borders.

## 2.6 Autonomous generation and discovery of cycles

Figure 8 demonstrates the agent's progressive interaction with the environment to autonomously generate Heat Pump Cycle No. 5 and Heat Engine Cycle No. 13. Rather than generating complete cycles in a single step, the agent incrementally builds structures by adding nodes and connections, transitioning from local linkages to globally closed loops. This evolutionary trajectory is consistent across both types of cycles.

Distinct construction strategies are also observed. Heat pump cycles exhibit multi-branch parallel development, with multiple potential closure paths emerging at intermediate stages. In contrast, heat engine cycles tend to first establish the main high- and low-pressure loops, followed by the addition of auxiliary branches. This difference is attributable to cycle characteristics: in a heat pump cycle, the ejector is a compound pressure component, whereas in a heat engine cycle, all pressure components are simple.

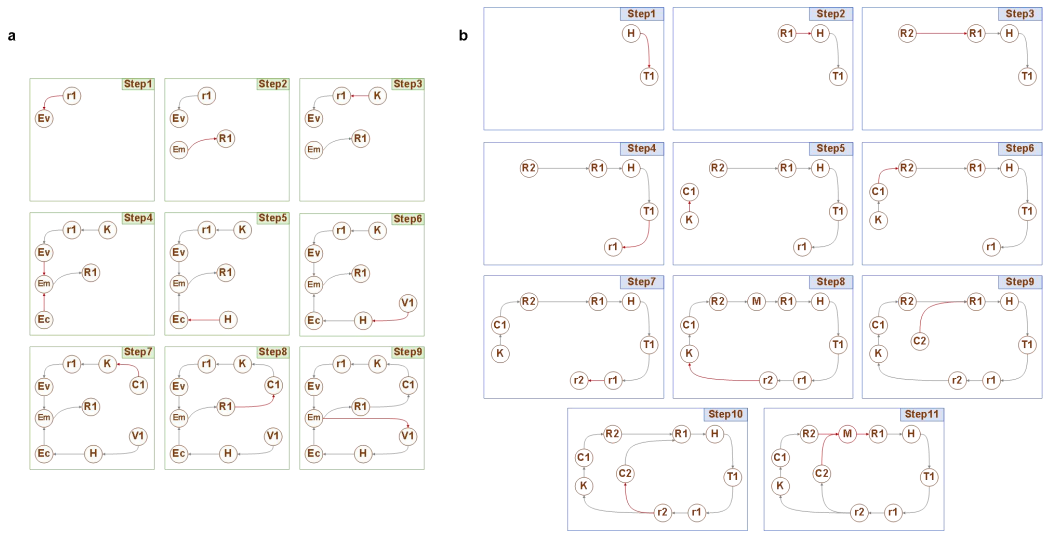

Figure 8 The process of the agent autonomously constructing cycles through stepwise interaction with the environment. a, Heat pump cycle No. 5; b, heat engine cycle No. 13.

Beyond autonomous generation, the agent also discovers numerous previously unreported cycle structures for both heat pump and heat engine systems. Among the 22 heat pump cycles, Nos. 1, 2, 3, and 5 correspond to established or expert-designed configurations[40, 41, 42] and are highlighted in Figure 5 with a light grey background, while the remaining 18 are newly discovered and marked with green borders. For the 26 heat engine cycles, Nos. 1, 2, 3, 8, and 13 are known configurations[43, 44] and are similarly highlighted in Figure 7, whereas the other 21 are newly identified and marked with blue borders. In terms of performance, some of these new cycles achieve or even exceed the performance of existing cycles, highlighting strong engineering potential.

# 3 Discussion

This study presents a novel framework integrating graph-based cycle encoding, physics-constrained graph decoding, and hierarchical reinforcement learning, establishing a new paradigm for intelligent thermodynamic system design. The Manager agent generates cycle structures that satisfy graph constraints, while the Worker optimizes operating parameters to maximize performance. Their collaboration enables the efficient exploration of complex design spaces, facilitating the autonomous generation and discovery of high-performance thermodynamic cycles. This surpasses the limitations of conventional enumeration and expert-driven approaches.

Optimization outcomes reveal key thermodynamic design principles. In particular, internal energy recovery mechanisms play a key role in enhancing thermodynamic performance. Many high-performance cycles incorporate the IHX components, which raise the working fluid's temperature before compression or heating, thereby reducing external energy input. Although this phenomenon aligns with classical thermodynamic principles, the agent's autonomous identification of these principles validates the framework's capacity to uncover underlying physical design rules within complex structural spaces.

While some heat engine cycles exhibit component redundancy under optimal conditions, transcritical $CO_2$ heat pump cycles demonstrate significant thermodynamic sensitivity to structural changes near critical points. Furthermore, although heat pump and heat engine cycles exhibit different patterns of performance improvement, the proposed framework consistently identifies high-performance cycle structures for both cases and can automatically adjust structural preferences with changing conditions. This suggests that, with appropriate graph rules, the framework can be extended to other complex thermodynamic systems.

Although the framework proposed in this paper exhibits strong capabilities in cycle design optimization, several areas warrant further investigation. Firstly, to ensure the physical feasibility of generated cycle structures, graph rules informed by domain knowledge were integrated during graph encoding and employed to evaluate agent actions within the RL environment. This method effectively curtails the search for invalid structures; however, the rules themselves remain partially manually defined. Future research could explore more general graph rule representations, enabling the agent to autonomously extract and form rules, thereby diminishing reliance on expert knowledge and enhancing generalization.

Secondly, the framework currently considers the COP of heat pump cycles and the thermal efficiency of heat engine cycles as singular optimization objectives. While this effectively identifies cycles with superior thermodynamic performance, practical engineering design also requires consideration of factors such as equipment count, system complexity, and economic cost. Future studies could extend the framework to multi-objective optimization.

Lastly, this study validated the method solely on heat pump and heat engine cycles.

Future research could apply the framework to more complex energy system design problems, such as combined cycle or integrated energy systems.

Overall, the proposed graph-based HRL framework offers a robust intelligent approach for automated co-design of thermodynamic cycle structures and parameters. Future work will focus on optimizing graph rules, incorporating multi-objective optimization, and extending the framework to more complex energy systems.

## 4 Methods

### 4.1 Deep Learning Neural Networks

A multi-layer perceptron (MLP), also known as an artificial neural network (ANN)[45], is employed as a regression model comprising multiple perceptrons. In Figure 1, the MLP surrogate predicts temperature, entropy, and dryness as outputs from pressure and enthalpy inputs. The input layer contains two nodes (pressure and enthalpy), while the output layer has three nodes (temperature, entropy, and dryness). Fully connected hidden layers propagate outputs from one layer to the next, with parameter propagation mathematically expressed in Equation (1). Here, $H_i$ denotes the output of the $i$-th hidden layer; $X$, the model input; $O$, the model output; $W_i$, the weights; $b_i$, the biases; and $\sigma$, the activation function. Activation functions employ the rectified linear unit (ReLU), defined in Equation (2).

Training data for the thermophysical property surrogate were generated using the open-source CoolProp library. The networks were trained using the Adam optimizer[46] for up to 500 epochs, incorporating an early stopping patience of 20 epochs. Learning rates were dynamically adjusted with ReduceLROnPlateau, starting at $1\times10^{-3}$ and halving if validation loss plateaued for 5 epochs. Architectures for the four deep learning models (PH2TSQ, PS2H, P2TH_SAT, T2P_SAT) comprised hidden layers of sizes [256, 256, 128], [256, 256, 128], [256, 256, 128], and [128, 128, 64], respectively.

$$\begin{cases} H_1 = \sigma(XW_0 + b_0) \\ H_2 = \sigma(H_1 W_1 + b_1) \\ O = H_n W_n + b_n \end{cases} \quad (1)$$

$$\operatorname{ReLU}(x) = \max(x, 0) \quad (2)$$

### 4.2 Reinforcement Learning Method

The interaction between agent and the environment in RL is formulated as a Markov Decision Process (MDP)[47], with state transition probabilities as shown in Equation (3). The state $s_t$ is represented by the graph-encoded adjacency matrix; the action $a_t$ is defined as activating an edge ($a_{i,j} = 1$). Equation (4) represents a trajectory of the MDP under a given policy $\pi(a|s)$. $r_t$ is the immediate reward for a single-step action. A discount rate $\gamma \in (0,1]$ is introduced to reduce the weight of future returns, and the total return of the entire trajectory is calculated via Equation (5) and (6). The Manager aims to learn a policy maximizing expected returns, parameterized by $\theta$.

$$p(s_{t+1} | s_t, a_t, \cdots, s_0, a_0) = p(s_{t+1} | s_t, a_t) \tag{3}$$

$$\tau = s_0, a_0, s_1, r_1, a_1, \cdots, s_{T-1}, a_{T-1}, s_T, r_T \tag{4}$$

$$G(\tau) = \sum_{t=0}^{T-1} \gamma^t r_{t+1} \tag{5}$$

$$J(\theta) = \mathbb{E}_{\tau \sim p_\theta(\tau)}[G(\tau)] = \mathbb{E}_{\tau \sim p_\theta(\tau)}[\sum_{t=0}^{T-1} \gamma^t r_{t+1}] \tag{6}$$

For stable policy learning, the Manager employs the Proximal Policy Optimization (PPO) algorithm[48, 49], which constrains policy updates to balance stability and exploration, suitable for the sensitivity of cycle search problems. The agent uses an Actor-Critic[50, 51] architecture, trained with the Adam optimizer (initial learning rate 6×10⁻⁴), discount factor $\gamma$ =0.99, clipping coefficient $\varepsilon$ = 0.2, and value function weight 0.6. Entropy regularization enhances exploration during training.

**4.3 Simultaneous Solution and Optimization of Graph Node Parameters**

Activated nodes are identified from the graph, with state points created for each inlet and outlet, each characterized by unknown variables pressure (*p*), enthalpy (*h*), and mass flow (*m*). Equation (7) formulates the nonlinear relations of inlet and outlet parameters within each component based on its physical model, whereas Equation (8) imposes nonlinear equality constraints linking parameters across connected components. The MINPACK hybr algorithm[52] solves these equations, yielding the state point parameters, thereby decoding the graph.

$$p_{i,\text{out}}, h_{i,\text{out}}, m_{i,\text{out}} = f(p_{i,\text{in}}, h_{i,\text{in}}, m_{i,\text{in}}) \tag{7}$$

$$[p_{j,\text{in}}, h_{j,\text{in}}, m_{j,\text{in}}] = [p_{i,\text{out}}, h_{i,\text{out}}, m_{i,\text{out}}] \quad \text{if } a_{i,j} = 1 \tag{8}$$

The Worker's parameter optimization is formulated as a constrained optimization problem, expressed in Equation (9). The objective function is defined as the system performance coefficient (COP for heat pump or efficiency for heat engine). Optimization variables consist of two types: $x_n$ representing graph node parameters (*p*, *h*, and *m*) and $x_m$ representing system operating parameters to be optimized, such as compressor suction/discharge pressures or flow split ratios. Equality constraints $h(x_n, x_m) = 0$ correspond to Equations (7) and (8), while inequality constraints $g(x_n, x_m) \leq 0$ ensure satisfaction of physical limits, such as minimum heat exchanger temperature differences and compressor pressure/temperature limits.

$$\begin{cases} \text{optimization problem} & \min -f(x) \\ \text{objective function} & f(x) = \text{COP (for Heat Pump)} \\ & \quad\quad\quad\text{or } \eta_t \text{(for Heat Engine)} \\ \text{optimization variables} & x = [x_n, x_m] \\ & x_n = [p_i, h_i, m_i] \\ & x_m = [p_{dis}, p_{suc}, r_{split}, \cdots] \\ \text{equality constraints} & h(x_n, x_m) = 0 \\ \text{inequality constraints} & g(x_n, x_m) \leq 0 \end{cases} \quad (9)$$

Combined with the graph decoding process, a nested optimization approach is adopted. The inner loop solves the equality constraints to decode each state point and compute the objective function, while the outer loop optimizes $x_m$ to maximize the objective. Bayesian optimization[53] is employed for the outer loop, as illustrated in Figure 1. An Upper Confidence Bound (UCB) acquisition function is used, with 100 initial sampling points and 200 optimization iterations. This surrogate-assisted method efficiently navigates the parameter space despite computational expense.

**4.4 Manager-Worker Joint Training Strategy**

To mitigate uneven credit assignment and convergence issues in Manager-Worker training, three strategies are integrated: performance feedback backpropagation, elite cycle memory, and staged training.

First, the performance metrics obtained from the Worker are backpropagated through the trajectory, distributing rewards to each step of the current cycle generation process. Equation (10) defines the reward function with performance backpropagation, ensuring that the Manager's decisions at every step reflect the final cycle performance. $\alpha$ is the backpropagation weight, set to 0.9.

$$\begin{aligned} r_t &\leftarrow r_t + \alpha \cdot \gamma^{T-t} \cdot \text{COP} \\ \text{or} \quad r_t &\leftarrow r_t + \alpha \cdot \gamma^{T-t} \cdot \eta_t \end{aligned} \quad (10)$$

Second, to prevent high-performance cycles from being forgotten during subsequent training, an elite cycle memory retains top-performing cycles, ranked by performance, forming a set of K distinct elite trajectories. During Manager policy updates, a soft regularization term based on elite state-action pairs encourages reproduction of high-quality cycles without compromising on-policy learning stability. Equation (11) defines the negative log-likelihood constraint imposed on the current policy, with $p_\varepsilon$ representing the empirical distribution of the elite set. Equation (12) shows the updated policy loss, where $\lambda_{elite}$ is the elite memory weight.

$$loss_{elite} = \mathbb{E}_{a_t \sim p_\varepsilon}[-\log \pi_\theta(a_t | s_t)] \quad (11)$$

$$loss = loss_{PPO} + \lambda_{elite} loss_{elite} \quad (12)$$

Finally, staged training balances exploration and convergence: early stages use higher entropy weight (0.1) and lower elite memory weight (0.01) to encourage exploration; later stages reduce entropy weight (0.01) and increase elite memory weight (0.1) to focus on high-performance cycles.

By integrating performance-based reward backpropagation, elite cycle memory, and staged training, the agent achieves a synergy between performance-driven cycle optimization and long-term memory retention.

# References

1. Creutzig F, Breyer C, Hilaire J, Minx J, Peters GP, Socolow R. The mutual dependence of negative emission technologies and energy systems. *Energy & Environmental Science* **12**, 1805-1817 (2019).
2. DeAngelo J, *et al.* Energy systems in scenarios at net-zero CO2 emissions. *Nature Communications* **12**, (2021).
3. Mignone BK, *et al.* Drivers and implications of alternative routes to fuels decarbonization in net-zero energy systems. *Nature Communications* **15**, (2024).
4. Langevin J, Harris CB, Reyna JL. Assessing the Potential to Reduce U.S. Building CO2 Emissions 80% by 2050. *Joule* **3**, 2403-2424 (2019).
5. Zhang S, *et al.* Targeting net-zero emissions while advancing other sustainable development goals in China. *Nature Sustainability* **7**, (2024).
6. Plazas-Nino FA, Ortiz-Pimiento NR, Montes-Paez EG. National energy system optimization modelling for decarbonization pathways analysis: A systematic literature review. *Renewable & Sustainable Energy Reviews* **162**, (2022).
7. Rajan A, McKay IS, Yee SK. Continuous electrochemical refrigeration based on the Brayton cycle. *Nature Energy* **7**, 320-328 (2022).
8. Irwin L, Le Moullec Y. Turbines can use CO2 to cut CO2. *Science* **356**, 805-806 (2017).
9. Wang K, He YL, Zhu HH. Integration between supercritical CO2 Brayton cycles and molten salt solar power towers: A review and a comprehensive comparison of different cycle layouts. *Applied Energy* **195**, 819-836 (2017).
10. Lecompte S, Huisseune H, van den Broek M, Vanslambrouck B, De Paepe M. Review of organic Rankine cycle (ORC) architectures for waste heat recovery. *Renewable & Sustainable Energy Reviews* **47**, 448-461 (2015).
11. Ahmadi M, Zirak S. 3E analysis of sCO2 recuperator cycle with multi effect desalination and organic Rankine cycle to enhance environmental sustainability. *Scientific Reports* **15**, (2025).
12. Dai YP, Wang JF, Gao L. Parametric optimization and comparative study of organic Rankine cycle (ORC) for low grade waste heat recovery. *Energy Conversion and*

*Management* **50**, 576-582 (2009).

13. Yu BB, Yang JY, Wang DD, Shi JY, Chen JP. An updated review of recent advances on modified technologies in transcritical CO2 refrigeration cycle. *Energy* **189**, (2019).

14. Li WQ, Yue B, Zhang H, Zheng CY, Jiang PX, Zhu YH. Optimization of trans-critical CO2 high-temperature heat pump cycle and study of maximum heating temperature. *International Journal of Refrigeration* **177**, 99-110 (2025).

15. Esposito M, Lindenberg K, Van den Broeck C. Universality of Efficiency at Maximum Power. *Physical Review Letters* **102**, (2009).

16. Bryant SJ, Machta BB. Energy dissipation bounds for autonomous thermodynamic cycles. *Proceedings of the National Academy of Sciences of the United States of America* **117**, 3478-3483 (2020).

17. Linnhoff B, Hindmarsh E. THE PINCH DESIGN METHOD FOR HEAT-EXCHANGER NETWORKS. *Chemical Engineering Science* **38**, 745-763 (1983).

18. Yeomans H, Grossmann IE. A systematic modeling framework of superstructure optimization in process synthesis. *Computers & Chemical Engineering* **23**, 709-731 (1999).

19. Newman MEJ. The structure and function of complex networks. *Siam Review* **45**, 167-256 (2003).

20. Gao L, Cao T, Hwang Y, Radermacher R. Graph-based configuration optimization for S-CO2 power generation systems. *Energy Conversion and Management* **244**, (2021).

21. Cui MD, Wang BL, Wang CL, Wei FL, Shi WX. GraPHsep: An integrated construction method of vapor compression cycle and heat exchanger network. *Energy Conversion and Management* **277**, (2023).

22. Zhao DP, Deng S, Zhao L, Xu WC, Zhao RK, Wang W. From 1 to N: A computer-aided case study of thermodynamic cycle construction based on thermodynamic process combination. *Energy* **210**, (2020).

23. Sutton RS, Barto AG, Sutton RS, Barto AG. *Reinforcement Learning: An Introduction second edition Introduction* (2018).

24. Sutton RS, McAllester D, Singh S, Mansour Y. Policy gradient methods for reinforcement learning with function approximation. In: *13th Annual Conference on Neural Information Processing Systems (NIPS)*) (1999).

25. Mnih V*, et al.* Human-level control through deep reinforcement learning. *Nature* **518**, 529-533 (2015).

26. Vinyals O*, et al.* Grandmaster level in StarCraft II using multi-agent reinforcement learning. *Nature* **575**, 350-+ (2019).

27. Silver D*, et al.* Mastering the game of Go with deep neural networks and tree search. *Nature* **529**, 484-+ (2016).

28. Yang T, Zhao LY, Li W, Zomaya AY. Reinforcement learning in sustainable energy and

electric systems: a survey. *Annual Reviews in Control* **49**, 145-163 (2020).

29. Wang Y, Wu JD, He HW, Wei ZB, Sun FC. Data-driven energy management for electric vehicles using offline reinforcement learning. *Nature Communications* **16**, (2025).

30. Du Y*, et al.* Intelligent multi-zone residential HVAC control strategy based on deep reinforcement learning. *Applied Energy* **281**, (2021).

31. Franzoso A, Fambri G, Badami M. Deep reinforcement learning as a tool for the analysis and optimization of energy flows in multi-energy systems. *Energy Conversion and Management* **341**, (2025).

32. Karniadakis GE, Kevrekidis IG, Lu L, Perdikaris P, Wang S, Yang L. Physics-informed machine learning. *Nature Reviews Physics* **3**, 422-440 (2021).

33. Dulac-Arnold G*, et al.* Challenges of real-world reinforcement learning: definitions, benchmarks and analysis. *Machine Learning* **110**, 2419-2468 (2021).

34. Bacon PL, Harb J, Precup D, Aaai. The Option-Critic Architecture. In: *31st AAAI Conference on Artificial Intelligence*) (2017).

35. Pateria S, Subagdja B, Tan AH, Quek C. Hierarchical Reinforcement Learning: A Comprehensive Survey. *Acm Computing Surveys* **54**, (2021).

36. Nachum O, Gu SX, Lee H, Levine S. Data-Efficient Hierarchical Reinforcement Learning. In: *32nd Conference on Neural Information Processing Systems (NIPS)*) (2018).

37. Bell IH, Wronski J, Quoilin S, Lemort V. Pure and Pseudo-pure Fluid Thermophysical Property Evaluation and the Open-Source Thermophysical Property Library CoolProp. *Industrial & Engineering Chemistry Research* **53**, 2498-2508 (2014).

38. Huber ML, Lemmon EW, Bell IH, McLinden MO. The NIST REFPROP Database for Highly Accurate Properties of Industrially Important Fluids. *Industrial & Engineering Chemistry Research* **61**, 15449-15472 (2022).

39. Karniadakis GE, Kevrekidis IG, Lu L, Perdikaris P, Wang SF, Yang L. Physics-informed machine learning. *Nature Reviews Physics* **3**, 422-440 (2021).

40. Ma YT, Liu ZY, Tian H. A review of transcritical carbon dioxide heat pump and refrigeration cycles. *Energy* **55**, 156-172 (2013).

41. Austin BT, Sumathy K. Transcritical carbon dioxide heat pump systems: A review. *Renewable & Sustainable Energy Reviews* **15**, 4013-4029 (2011).

42. Song YL, Cui C, Yin X, Cao F. Advanced development and application of transcritical $CO_2$ refrigeration and heat pump technology-A review. *Energy Reports* **8**, 7840-7869 (2022).

43. Crespi F, Gavagnin G, Sánchez D, Martínez GS. Supercritical carbon dioxide cycles for power generation: A review. *Applied Energy* **195**, 152-183 (2017).

44. Liu YP, Wang Y, Huang DG. Supercritical CO2 Brayton cycle: A state-of-the-art review. *Energy* **189**, (2019).

45. Hornik K, Stinchcombe M, White H. Multilayer feedforward networks are universal

approximators. *Neural networks* **2**, 359-366 (1989).

46. Kingma DP, Ba J. Adam: A method for stochastic optimization. *arXiv preprint arXiv:14126980*, (2014).
47. Abounadi J, Bertsekas D, Borkar VS. Learning algorithms or Markov decision processes with average cost. *Siam Journal on Control and Optimization* **40**, 681-698 (2001).
48. Schulman J, Levine S, Moritz P, Jordan M, Abbeel P. Trust Region Policy Optimization. In: *32nd International Conference on Machine Learning*) (2015).
49. Schulman J, Wolski F, Dhariwal P, Radford A, Klimov O. Proximal policy optimization algorithms arXiv. *arXiv (USA)*, 12 pp.-12 pp. (2017).
50. Peters J, Schaal S. Natural Actor-Critic. *Neurocomputing* **71**, 1180-1190 (2008).
51. Bhatnagar S, Sutton RS, Ghavamzadeh M, Lee M. Natural actor-critic algorithms. *Automatica* **45**, 2471-2482 (2009).
52. Virtanen P*, et al.* SciPy 1.0: fundamental algorithms for scientific computing in Python. *Nature Methods* **17**, 261-272 (2020).
53. Wang X, Jin Y, Schmitt S, Olhofer M. Recent Advances in Bayesian Optimization. *Acm Computing Surveys* **55**, (2023).

## Acknowledgments

This work was supported by the National Science and Technology Major Project (Project No. 2026ZD1702400) and Fundamental and Interdisciplinary Disciplines Breakthrough Plan of the Ministry of Education of China (JYB2025XDXM304).

## Supplementary information

Table S1 Prediction error statistics of the deep learning surrogate models for thermophysical properties. The horizontal axis represents the relative error interval, and the vertical axis represents the proportion of samples. PH2TSQ predicts temperature, entropy, and vapor quality from pressure and enthalpy; PS2H predicts enthalpy from pressure and entropy; P2TH_SAT predicts saturated temperature and saturated liquid and vapor enthalpies from pressure; T2P_SAT predicts saturated pressure from temperature.

| Relative error range (%) | Proportion | | | | | | | |
|---|---|---|---|---|---|---|---|---|
| | PH2TSQ T | PH2TSQ S | PH2TSQ Q | PS2H H | P2TH_SAT T | P2TH_SAT HV | P2TH_SAT HL | T2P_SAT P |
| (0,0.01] | 0.531 | 0.551 | 0.070 | 0.315 | 0.923 | 0.933 | 0.668 | 0.650 |
| (0.01,0.05] | 0.447 | 0.430 | 0.256 | 0.584 | 0.077 | 0.067 | 0.271 | 0.311 |
| (0.05,0.1] | 0.020 | 0.017 | 0.198 | 0.091 | 0 | 0 | 0.050 | 0.030 |
| (0.1,0.5] | 0.002 | 0.002 | 0.329 | 0.010 | 0 | 0 | 0.012 | 0.009 |
| (0.5,1] | 0 | 0 | 0.144 | 0 | 0 | 0 | 0 | 0 |

| $(1,+\infty)$ | 0 | 0 | 0 | 0 | 0 | 0 | 0 | 0 |